# Robot Arm Control via Cognitive Map Learners

Nathan McDonald
Christian Brazeau
Air Force Research Laboratory
nathan.mcdonald.5@us.af.mil

Colyn Seeley
Rochester Institute of Technology

*Abstract*— Cognitive map learners (CML) have been shown to enable hierarchical, compositional machine learning. That is, independently trained CML modules can be arbitrarily composed together to solve more complex problems without task-specific retraining. This work applies this approach to control the movement of a multi-jointed robot arm, whereby each arm segment's angular position is governed by an independently trained CML. Operating in a 2D Cartesian plane, target points are encoded as phasor hypervectors according to fractional power encoding (FPE). This phasor hypervector is then factorized into a set of arm segment angles either via a resonator network or a modern Hopfield network. These arm segment angles are subsequently fed to their respective arm segment CMLs, which reposition the robot arm to the target point without the use of inverse kinematic equations. This work presents both a general solution for both a 2D robot arm with an arbitrary number of arm segments and a particular solution for a 3D arm with a single rotating base.

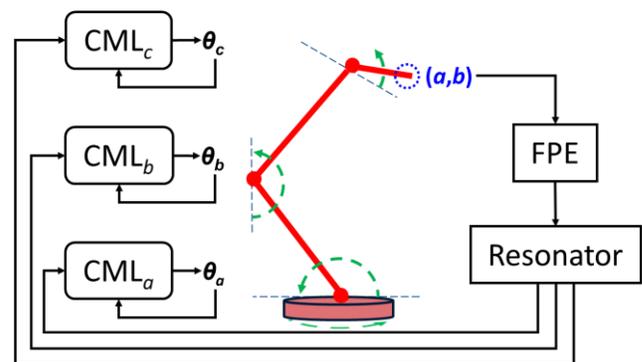

Fig. 1. Robotic arm with 3 arm segments, articulated by 3 different angles (green dashed lines), target point (*a*, *b*), and workflow diagram for repositioning the robot arm. Target point is converted to a phasor hypervector via fractional power encoding (FPE), which is then factorized by the resonator into a set of arm segment angles received as input by their respective CML. CMLs plot sequence of angular states in parallel.

## I. INTRODUCTION

Deep neural networks (DNN) are frequently implemented as monolithic solutions, e.g. a single DNN trained for a single task. A limitation of this approach is that training by stochastic gradient precludes trivial post-training composability of diverse pretrained DNNs. Even ensemble networks deliberately constructed with multiple pretrained DNNs must still be fine-tuned as a monolithic entity for optimal results [1]. DNNs are therefore not suitable as the primary module for composable machine learning. Such a framework might consist of 1) fundamental computing modules that 2) may be arbitrarily composed together to 3) solve more complex operations than the modules individually yet 4) without reoptimizing the modules per new task, typically necessitating 5) a consistent input/out (I/O) representation [2, 3]. (For example, digital logic is a composable framework where Boolean logic gates are the composable modules and binary the common I/O.)

This work presents an example of cognitive map learners (CML) [4] as composable ML modules operating upon hyperdimensional computing (HDC) hypervectors as the "common communication protocol" [3, 5, 6]. CMLs are a collection of separate single-layer artificial neural networks (matrices) collaboratively trained to learn internal representations of node states, edge actions, and edge action availability from an abstract graph. The key capability of this unusual ML module is that it iteratively performs path planning between any current and desired node state pairing, though never explicitly trained to do so. CMLs can be readily trained or even explicitly constructed [6] to learn discrete state spaces; however, training a CML on continuous state spaces often require numerous attempts to find a viable solution, e.g. a state space comprised of the position angles of all four legs of a quadruped robot [4].

Instead of attempting to learn this continuous state space directly, this work describes a preliminary approach to control a single multi-jointed robot arm via a collection of discrete space CMLs: one CML per arm segment. A target point is defined in 2D Cartesian space, encoded as a phasor hypervector via fractional power encoding (FPE), then factorized into a set of arm angles also expressed as hypervectors from the CMLs. From there, the arm segment CMLs plot a path to the desired angles, repositioning the whole arm arbitrarily close to the target point (Figure 1). This approach is a general solution for manipulating a 2D robotics arm comprised of an arbitrary number of arm segments. Since rotation are not commutative, only a particular solution to 3D arm control is presented.

The main contributions of this work are as follows:

- Defined a workflow for generating target points and arm segment CMLs in a shared representation space via fractional power encoding (FPE) of phasor hypervectors

- Reinterpreted finding the set of arm angles attaining the target point as equivalent to the knapsack problem, solvable via hypervector factorization methods.

- Detailed a method for repositioning a robot arm via CMLs without inverse kinematics equations.

Section II describes the mathematics of HDC and CMLs, while Section III details both the construction of the arm segment CMLs and the workflow for their orchestration. Section IV presents the results for this approach, followed by discussion and future applications of this research in Section V.

## II. BACKGROUND

### A. Cognitive Map Learner (CML)

A cognitive map learner is a collection of three single layer neural networks (matrices) which are trained or constructed to encode the topology of an abstract graph of $n$ nodes and $e$ edges. Each neural network learns one aspect of the graph: 1) node state representations $S \in \mathbb{R}^{(d,n)}$, 2) edge action representations $A \in \mathbb{R}^{(d,e)}$, and 3) the availability $G \in \mathbb{R}^{(e,n)}$ of edge actions from each node state, where $d$ is the size of the representation space. A CML is trained to predict the next node state $s_{t+1}$ given the current node state $s_t$ and chosen edge action $a_t$,

$$s_{t+1} = s_t + a_t, \quad (1)$$

For the use case here, the node states $S$ are predefined by fractional power encoding with phasor hypervectors (Section III. A, B), so the edge action matrix $A$ is calculated explicitly as

$$a_{ij} = s_i - s_j. \quad (2)$$

The chief use of the CML is for path planning, despite not having been explicitly trained for the task (only pairwise state transitions are presented in Eq. (2)). The CML module takes two inputs: the current node state $s_t$ and a desired node state $s_d$ (Figure 2). Because all node states are unambiguously stored as column vectors in $S$, they may be trivially extracted from any pretrained CML, providing the vocabulary for an external agent to task the CML. The CML can also clean up noisy inputs $\hat{s}$ provided they exceed a minimum similarity threshold with respect to the known state hypervectors in $S$, Eq. (6).

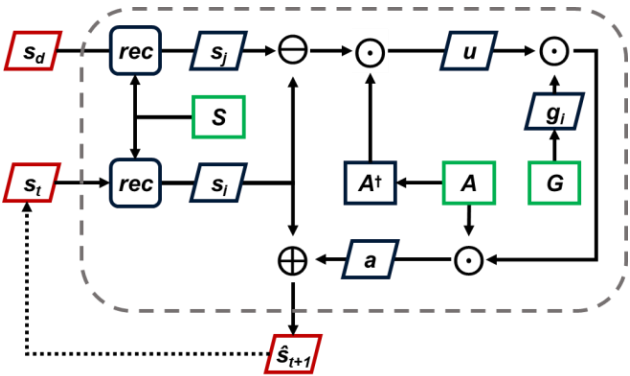

Fig. 2. CML wiring diagram: inputs $s_d$ and $s_t$; output $\hat{s}_{t+1}$; and internal node state representations $S$, edge actions $A$, and gating $G$.

For each iteration of the CML, the utility of every edge action with respect to $s_d$ is calculated by multiplying the difference between the desired and current node state by the pseudo-inverse of edge action matrix $A$,

$$u = A^\dagger (s_d - s_t), \quad (3)$$

where $\dagger$ denotes the Moore-Penrose pseudo-inverse. The gating vector $g_t$ (corresponding to node $s_t$) is multiplied elementwise with the utility scores vector, ensuring only legal edge actions have nonzero values. A winner-take-all (*WTA*) algorithm produces a one-hot vector indicating the index of the most useful edge action, which is then added to the current node state

$$\hat{s}_{t+1} = s_t + A \cdot WTA(g_t \odot u), \quad (4)$$

where $\odot$ denotes elementwise multiplication; and the CML returns $\hat{s}_{t+1}$ as output, the predicted most useful, legal next node state. Since this output is also a valid input, the CML can iterate over Eq. (3, 4) until $s_d \approx s_t$; whereby, the CML finds a reasonably minimal path between any initial and desired node state, competitive with Dijkstra and A* [7] but without the mathematical optimality guarantees [4].

### B. Hyperdimensional Computing (HDC)

By making the CML representation space size $d \geq 512$, node and edge vectors become semantically meaningful entities suitable for symbolic reasoning using hyperdimensional computing (HDC) algebra [8, 9]. The key metric under HDC is similarity *sim*. As the length of randomly generated hypervectors increases, the similarity between any two converges to pseudo-orthogonality [10]. The basic operations of HDC, viz. addition, multiplication, and recovery, therefore create and manipulate similarities among hypervectors. Addition and multiplication are elementwise operations, so the dimension of the resultant hypervector remains $d$ regardless of the number of hypervectors added or multiplied together. The normalized dot product is used to compare hypervectors, where identical vectors have a similarity of *sim* = 1 and pseudo-orthogonal hypervectors *sim* ~ 0.

Addition is analogous to set creation, where the sum hypervector is similar to each of the component hypervectors,

$$q = [s_1 + s_2 + s_3], \quad (5)$$

and $sim(q,s_1) \sim sim(q,s_2) \sim sim(q,s_3) \gg sim(q,s_4) \sim 0$. Multiplication, denoted as $\odot$, binds hypervectors together, analogous to key-value pairing. Unlike addition, the product hypervector is not similar to either of its factor hypervectors, $sim(x \odot y, x) \sim 0$; but importantly the operation is reversible (Section III. A). The recovery, or cleanup, operation *rec* compares a noisy query hypervector to all other known hypervectors stored in a dictionary and returns the most similar hypervector above a threshold $\theta$,

$$rec(\hat{s}_1, S, \theta) = s_1. \quad (6)$$

## III. METHODS

### A. Specifying a target point

Let a robotic arm have $n$ arm segments. The goal is to determine the angle of each arm segment such that the end effector of the arm touches a target point (Figure 1). Cartesian axes $x$ and $y$ are each defined as a unique base vector of length $d = 1024$, whose elements are phasors, complex values with unit amplitude,

$$x = e^{j2\pi U}$$
$$y = e^{j2\pi U}, \quad (7)$$

where $j$ is the imaginary unit and $U$ is a uniform random vector with elements $[0,1]$. Given target point $(a, b) \in \mathbb{R}$, each hypervector $x$ and $y$ is raised to the corresponding power, hence the name fractional power encoding (FPE) [11], then multiplied together,

$$(a, b) \Rightarrow p = x^a \odot y^b, \quad (8)$$

What makes FPE uniquely suitable for this application is that multiplication of phasors is equivalent to addition of their exponents (Figure 3). Encoded positions can be further manipulated along both axes by subsequent multiplication, including negative exponents (multiplicative inverse) through use of the complex conjugate *, e.g.

$$\begin{aligned} x^3 y^4 &= (x^1 y^{2.2}) \odot (x^2 y^{1.8}) \\ &= (x^5 y^5) \odot (x^2 y^1)^*, \end{aligned} \quad (9)$$

where the $\odot$ between $x$ and $y$ is dropped for visual clarity.

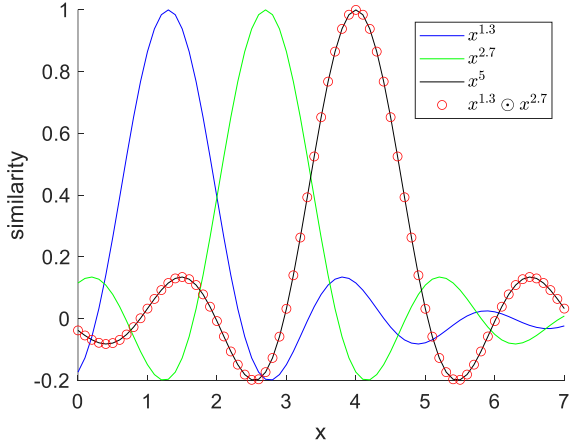

Fig. 3. Similarity of FPE encoded points, illustrating multiplication as addition of exponents.

### B. Arm segment CML

Consider a single arm segment of length $r$ about a pivot, tracing a circle in the 2D plane. The accessible angle states per arm segment are restricted as the $m^{\text{th}}$ roots of unity,

$$\omega_k = 2\pi k/m, \quad (10)$$

where $k = \{1, 2, \ldots, m-1, m\}$ (Figure 4).

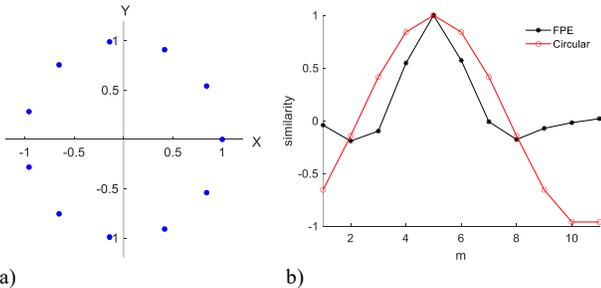

a)          b)

Fig. 4. a) $m = 11$ discrete arm segment angle states (blue dots) and b) similarities among node states phasor hypervectors using FPE and circular encoding.

To interact with the target point defined in Eq. (8), each of the $m$ node states $s_m$ are also encoded via FPE using the same $x$ and $y$ basis vectors from Eq. (7),

$$\begin{aligned} a &= r \cos(2\pi k/m) \\ b &= r \sin(2\pi k/m) \end{aligned} \quad (11)$$
$$s_k = x^a \odot y^b. \quad (12)$$

In the first 2D case, let the arm segments not be constrained by physics, namely permitted to pass through one another. In this way, each arm segment CML learns a bidirectional ring graph. Since each angle state $s_k$ is precomputed by Eq. (12), each edge action $a$ between sequential states $s_{k+1}$ and $s_k$ is explicitly calculated as

$$a_{k+1,k} = s_{k+1} - s_k. \quad (13)$$
$$a_{k,k+1} = s_k - s_{k+1}. \quad (14)$$

A CML $C_n$ is created for each arm segment, and each $C_n$ is created independently of any other CML, excepting that all CMLs must use the same $x$ and $y$ basis vectors, Eq (7).

For a concrete baseline, let $r = 1$ for $n = 3$ arm segments. Let each arm segment touch $m$ points on the circumferences of the unit circle. Given a sufficiently high number of states $m$ per arm segment, any point satisfying $3^2 \geq x^2 + y^2$ is expected to be a valid robot arm effector position. Therefore, given target hypervector $p$ from Eq. (8), the goal is to recover sets of arm position states $s_m$ of the form

$$\begin{aligned} p &= x^a y^b \\ &\approx x^{(i+j+k)} y^{(l+m+n)} \\ &\approx x^i y^l \odot x^j y^m \odot x^k y^n, \end{aligned} \quad (15)$$

where the product of the recovered factors has a similarity greater than a threshold $\theta$.

### C. Factoring hypervectors

Recall that for HDC algebra, multiplication results in a product hypervector pseudo-orthogonal to its factors. Therefore, factoring a product hypervector via a similarity search alone results in a combinatorial explosion problem, since even partial products of factors are dissimilar. However, both a resonator network [12, 13] and a modern Hopfield network [14] can efficiently factor product hypervectors given codebooks of the candidate vectors.

For the $n = 3$ arm segments case, the three CML angle state matrices $S$ serve as codebooks of candidate factors, which for visual clarity are referred to here as $A = [a_1, \ldots, a_m]$, $B = [b_1, \ldots, b_m]$, and $C = [c_1, \ldots, c_m]$. Given target point $p \approx a_i \odot b_j \odot c_k$, to factor via a resonator network, the following resonator module equations are iterated

$$\hat{a}(t+1) = f(AA^\top (p \odot \hat{b}^*(t) \odot \hat{c}^*(t)))$$
$$\hat{b}(t+1) = f(BB^\top (p \odot \hat{a}^*(t) \odot \hat{c}^*(t)))$$
$$\hat{c}(t+1) = f(CC^\top (p \odot \hat{a}^*(t) \odot \hat{b}^*(t))), \quad (16)$$

where $\hat{a}$, $\hat{b}$, and $\hat{c}$ are estimations of a factor from each codebook, * denotes the complex conjugate, and $f$ forces the magnitude of all elements to 1. The current arm angle estimates $\hat{a}$, $\hat{b}$, and $\hat{c}$ are initialized as some combination of 1) the current arm segment state, 2) the sum (or superposition) of all node states of the respective codebook, and/or 3) a random phasor hypervector to inject noise into the system. Every iteration $t$, each resonator module recovers an estimate of its own factor by unbinding the estimates from all other modules with the target point. The decoded hypervector is then compared to the module's codebook, and an auto-associative memory attempts to converge each estimate. In this way, the resonator network leverages superposition to search multiple potential factorizations simultaneously. Each new factor estimate is then shared with the other modules to update their own factor estimates.

After every 10 iterations, the similarity of each estimated factor is measured against its respective codebook. If the index of the most similar entry is the same for the last two iterations per codebook, then the resonator algorithm terminates and the actual codebook entries are returned (Figure 5a). Note, it is not uncommon for the resonator to converge on pairs of anti-parallel versions of the factor, $sim(a,-a) = -1$. Mathematically, this is because negatives cancel out, $sim(a \odot b, -a \odot -b)$. The proposed arm configuration to reach the target point is computed directly from the indexed codebooks entries,

$$\hat{p} = a_i \odot b_j \odot c_k. \qquad (17)$$

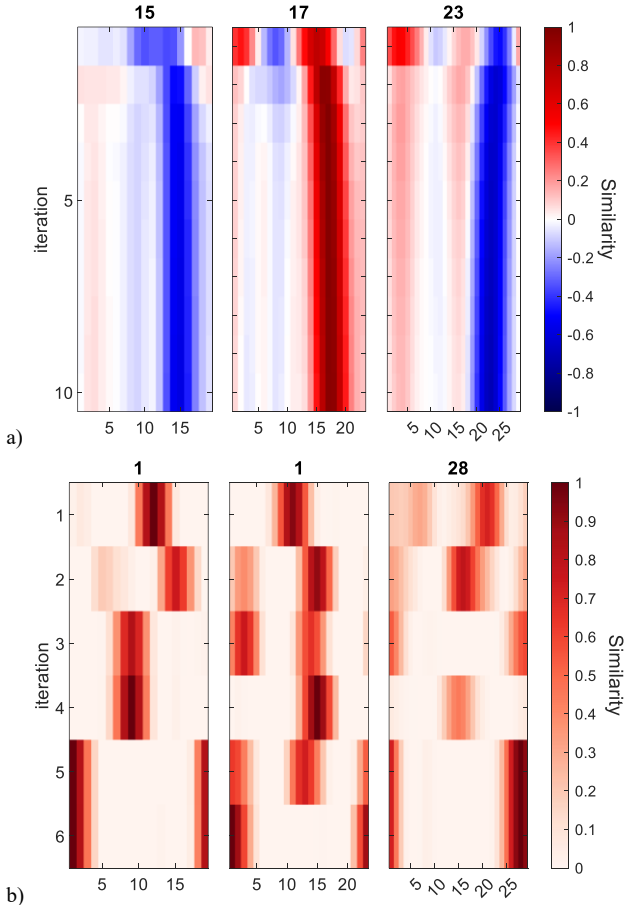

a)
b)
Fig. 5. a) Factorization via resonator network, where each row is one iteration of the network. Colors indicate the similarity of the estimated factors with respect to their codebooks. Figure shows convergence to anti-parallel versions of codebook entries. b) Factorization via modern Hopfield network.

This is because the resonator only approximates the recovery operation on the estimated factors, Eq. (6). Figure 8a shows the range of estimated factor similarities when the resonator algorithm terminates. However, the *recovery* operation can be subsequently applied successfully to all estimated factors, since their similarities are all greater than the noise floor for random vectors, $\eta_{floor} = 0.05$ for $d = 1024$.

Alternatively, a modern Hopfield network is used [14]. While the original Hopfield network is an iterative algorithm, minimizing the energy calculation between the input state pattern and that stored in the network, this updated algorithm attains near optimal pattern restoration after only a single iteration,

$$s = softmax(\beta S^\top \hat{s})S, \qquad (18)$$

where $S$ is the codebook, $\hat{s}$ is the initial query vector, $s$ is the recovered pattern, and $\beta$ is a bias term. Given target point $p$ as before, to factor via a modern Hopfield network, the following equations are iterated

$$\hat{a} = softmax(\beta A^\top (p \odot \hat{b}^* \odot \hat{c}^*))A$$
$$\hat{b} = softmax(\beta B^\top (p \odot \hat{a}^* \odot \hat{c}^*))B$$
$$\hat{c} = softmax(\beta C^\top (p \odot \hat{a}^* \odot \hat{b}^*))C. \qquad (19)$$

After each iteration, the similarities of the estimated factors are measured with their respective codebooks. Unlike a resonator network which converges to a solution, the Hopfield network jumps among solutions over subsequent iterations (Figure 5b). So for a termination condition, per iteration, if each estimated hypervector has a similarity greater than $sim_{max}(\hat{s}, S) \geq 0.75$, then Hopfield network is considered converged and the actual codebook entries are returned.

Regardless of factorization method, after 50 iterations, if either 1) not all the codebooks have converged or 2) the proposed arm configuration $\hat{p}$ is below the similarity threshold, then the factorization method is reinitialized, with a random subset of estimated factors initialized as random phasor hypervectors to inject noise into the system. If after 50 reinitializations an acceptable solution is still not found, then the threshold is reduced by 0.01 until the system succeeds.

### D. Repositioning arm segments via CMLs

Lastly, each arm segment CML receives as input the respective recovered factor $s_n = \{a_i, b_j, c_k\}$ as the desired state $s_d$. Each CML then iteratively plots a state transition path from its current angle state $s_t$ the desired state $s_d$ in parallel; whereupon the robot arm effector approaches the target point $(a, b)$ with arbitrary precision.

### E. 3D arm control

2D vector addition is commutative, and this property is preserved in the elementwise multiplication of FPE hypervectors, such that the robot end effector position is unaffected by the order of the arm segments. This is not true when operating in 3D space, since rotations are not commutative. The following is one possible solution for controlling an $n = 3$ arm segments arm with only a rotating base.

The target point $(a, b, c)$ in Cartesian space is first converted to spherical coordinates $(\rho, \theta, \phi)$. $\rho$ and $\theta$ are then re-converted back to the XY plane,

$$p = x^{\rho \cos\theta \cos\phi} y^{\rho \cos\theta \sin\phi}. \quad (20)$$

The above 2D arm positioning algorithm is then run as before to position the arm end effector to point $(\rho, 0, \phi)$.

Instead of using FPE to encode the final rotation of the arm to $\theta$, circular coding is used for the CML. The initial state of the rotating base is initialized as

$$s_1 = e^{j\frac{2\pi}{m}U}, \quad (21)$$

where $U$ is a uniform random distribution of integers $k = \{1, 2, \ldots, m\}$ and $m$ is the total number of states to be generated. Each subsequent state is incremented as

$$s_{k+1} = s_k \odot e^{j\frac{2\pi}{m}}. \quad (22)$$

That is, the FPE arm segment CML states encode allocentric orientation; whereas, the circular CML encodes egocentric orientation. The circular CML only has two edge action states: forward (counter-clockwise) and backwards (clockwise),

$$a_1 = e^{j\frac{2\pi}{m}}, \quad (23)$$

$$a_2 = e^{-j\frac{2\pi}{m}}, \quad (24)$$

respectively. Due to the periodic nature of phasors, after $m$ nodes, the node states repeat, $s_1 = s_m \odot a_1$ (Figure 5b). Since this circular CML shares no basis vectors with the arm segment CMLs, a dictionary must be maintained pairing the codebook index with a real-valued angle. Given target angle $\theta$, the index of the circular CML state closest to that angle is selected as the desired state $s_d$ for the circular CML. The circular CML thereby rotates the whole arm to arrive at $(\hat{\rho}, \hat{\theta}, \hat{\phi})$, approximating the target point $(a, b, c)$.

IV. RESULTS

A. 2D unconstrained arm

For the unconstrained 2D case, the arm segments are allowed to pass through one another. Given an arm of $n = 3$ arm segments of length $r = 1$, then all points satisfying $3^2 \geq x^2 + y^2$ are expected to be accessible by the robot arm. As a practical measure, this space was discretized into 0.1x0.1 grid cells, defining 2,809 possible target points $p$. When each arm had $m = 29$ angle positions, there were 24,389 possible end effector states (Figure 6a); however, only 74.5% of the aforementioned grid cells were reachable due to the highly structured point distribution (Figure 6b). However, leveraging the insights of residue number systems [15], giving the arm segment CMLs sequential co-primes of angle states $m = [19, 23, 29]$ covered 98.8% of the available grid cell points (Figure 6d) yet with only 12,673 end effector states due to a more diffuse point spread (Figure 6c).

While the target point was given in Cartesian space, the factorization routines were contingent upon only the similarity of the reconstructed product hypervector $\hat{p}$ and the target vector $p$, $sim(\hat{p}, p) \geq \theta$. Heuristically, a hypervector similarity threshold of $\theta \geq 0.99$ corresponded well with an arbitrarily acceptable

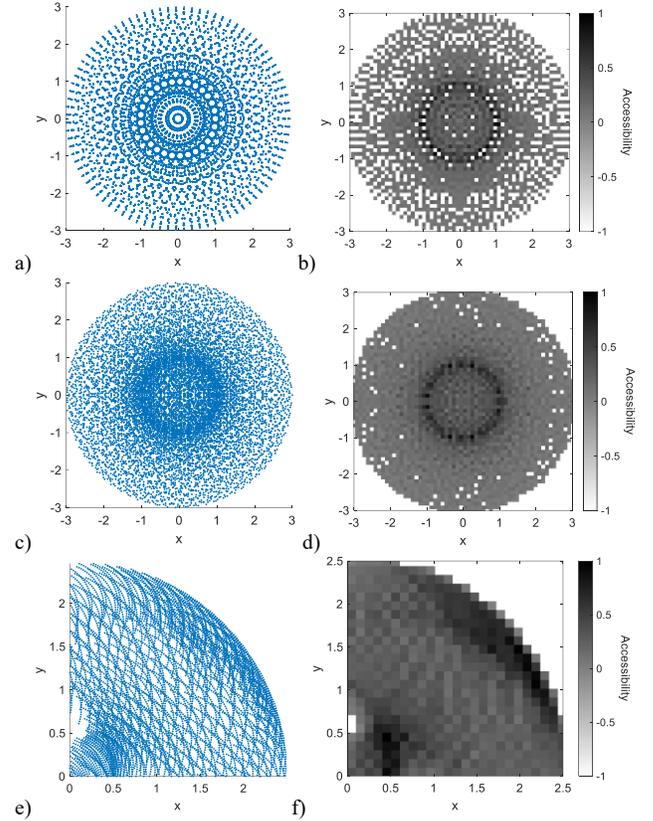

Fig. 6. For $m = [29, 29, 29]$, a) all 24,389 attainable points and b) corresponding histogram when mapped to 0.1×0.1 grid cells, covering only 74.5% of possible cells. For $m = [19, 23, 29]$, c) all 12,673 attainable points and d) histogram covering 98.8% of possible cells. e) For the gripper arm, all attainable points within the first quadrant and f) histogram covering 99.4% of cells. Inaccessible points (white cells) are color shifted to enhance contrast.

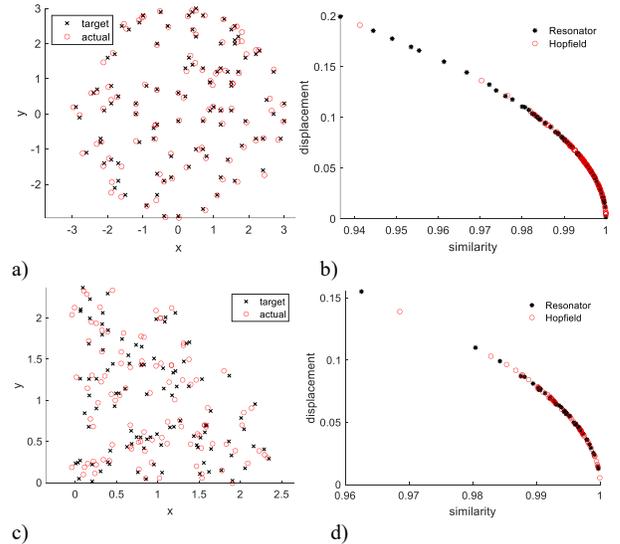

Fig. 7. Scatter plot of target point and final arm position for a) 2D and c) 3D arms. Strong correlation between Euclidean displacement and hypervector similarity for resonator network and Hopfield network for b) 2D and d) 3D arms.

Euclidean displacement of <0.1 (Figure 7a). The mean displacement using the resonator was $\epsilon = 0.071 \pm 0.036$ and $\epsilon =$

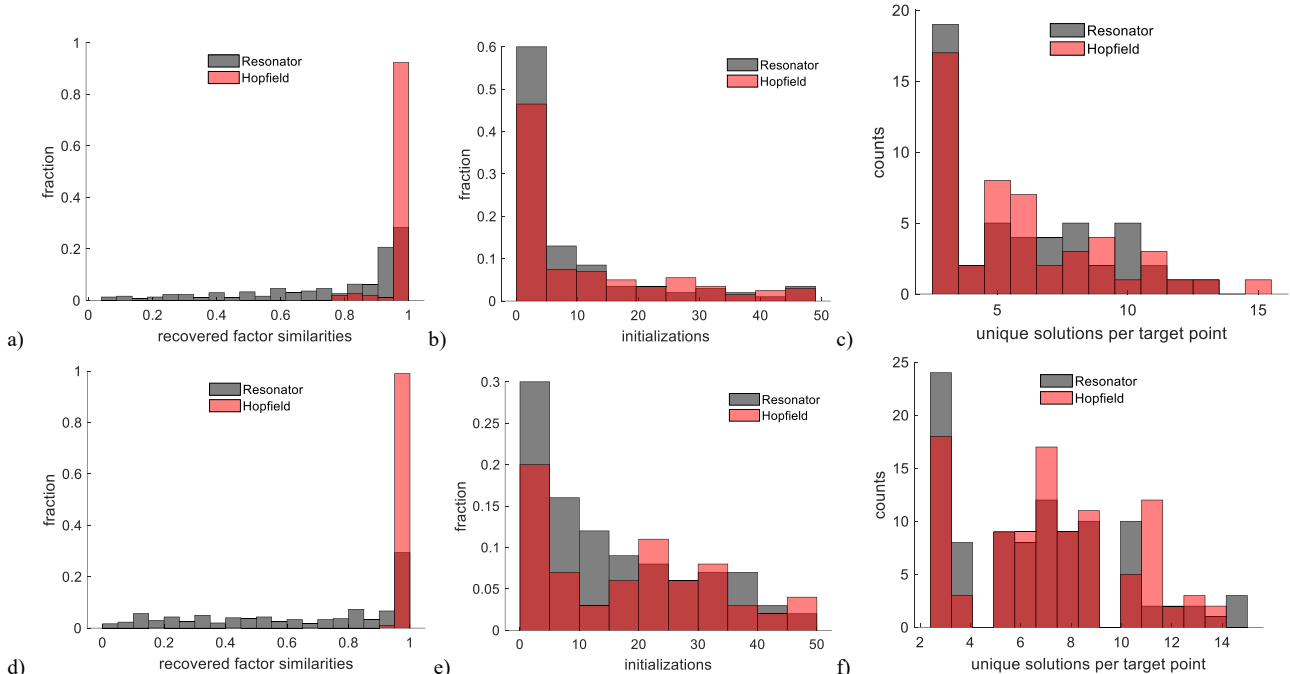

Fig. 8. Histogram of the similarities of the estimated factors with their codebook entry, $\delta(\hat{a}, a)$, upon completion of factorization for a) unconstrained 2D and d) constrained 3D. Histogram of factorization initializations to recover arm position above threshold for b) 2D and e) 3D. Number of unique arm positions for 100 points with 20 initializations for c) 2D and f) 3D.

$0.061 \pm 0.030$ using the Hopfield network. However, the high similarity threshold required multiple factorization initializations (Figure 8b). Out of 100 random accessible target grid cell points, the resonator network found a solution on the first initialization attempt 14% of the time (mean $15.0 \pm 12.1$ initializations) and the Hopfield network 10% of the time (mean $31.6 \pm 25.9$ initializations).

As shown in Figure 6d, the number of arm configurations that reached a grid cell point varied significantly. Since factorization is deterministic, the inclusion of random phasor hypervectors as part of initialization prompted the system to find multiple acceptable arm configurations. Given 100 random accessible grid cell points and 20 factorization initializations each, the resonator found $5.1 \pm 2.3$ arm configurations per point and the Hopfield network found $5.8 \pm 2.2$ (Figure 8c). Presently, there is no mechanism included for evaluating among multiple proposed solutions. A possible selection criterion is to minimize the total change across the arm angle states since the index the target and current node states are known.

### B. Constrained 3D arm

To illustrate the relevance of this technique to 3D physics-based robots, the $n = 3$ arm segments were not permitted cause any part of the arm to contact the floor nor pass through one another. These physical limitations were enforced through the arm segment CMLs directly. The arm segment graphs were made linear graphs (as opposed to ring graphs) by removing the bidirectional edges between the first $s_1$ and last $s_m$ node state. The last arm length was shorted to $r_c = 0.5$, and the maximum range of movement for each arm was restricted and then further offset (Figure 1):

$$\omega_a = [0, \pi],$$
$$\omega_b = [0, \pi] + 3\pi/2,$$
$$\omega_c = [0, \pi/2] + 15\pi/8. \quad (25)$$

Since the circular CML for $\theta$ can be made arbitrarily close, the following results focus on attaining $(\rho, \phi)$ in the 2D XY plane. These restrictions limited the arm to operate only in the first quadrant of the Cartesian space yet preserved 99.4% coverage (Figure 6e, f). As before, the mean displacement using the resonator was $\epsilon = 0.082 \pm 0.041$ and the Hopfield network was $\epsilon = 0.065 \pm 0.026$ (Figure 7c, d). Out of 100 random accessible target grid cell points, the resonator network found a solution on the first initialization attempt 15% of the time (mean $13.5 \pm 12.4$ initializations) and the Hopfield network 13% of the time (mean $35.1 \pm 29.9$ initializations) (Figure 8e). Lastly, given 100 random accessible grid cell points and 20 factorization initializations, the resonator found $6.74 \pm 3.22$ arm configurations per point, while the Hopfield network found $7.28 \pm 3.01$ (Figure 8f).

## V. DISCUSSION

The emphasis on this work was to illustrate how CMLs can be arranged to collaboratively control 2D and 3D robot arm movement. In essence, this approach reinterpreted arm positioning as a special case of the knapsack problem, a subset sum problem. Due to the superposition properties of HDC, factorization techniques exist that can efficiently solve this task. At this time, a formal comparison against standard inverse kinematics equation solvers has not been accomplished, which would include additional metrics such as computation time and path length [16, 17]. It is worth noting meanwhile that this CML framework avoids the singularity failure condition of inverse kinematics solvers by iteratively relaxing of the minimum similarity threshold until a solution is produced (Figure 7b, d). The current framework also does not include environmental sensors to invoke an obstacle avoidance routine, but it has already been shown that CMLs immediately adapt to changes in state or action availability [4]. Renner, *et al.* also described non-commutative factorization techniques which may be applicable to expanding the framework described here to control a standard 6-degree of freedom robot arm [18, 19]. The same work also described a method for implementing phasor HDC operations in spiking neural networks on the Loihi neurotrophic hardware, which could enable deployment of these algorithms on smaller robotic platforms.

The proposed 3D arm positioning method will next be applied to the inciting observation about the limitations of using a CML to control the quadruped robot. Because the arm segments states are created via FPE, it is straightforward to copy the CML framework, amend the node states $S$, and recompute the action matrix $A$ for each additional leg. Using a central pattern generator to prescribe target leg positions in time, the proposed method is expected to control a basic quadruped robot, though with nominally 12 CMLs (3 CMLs per leg) instead of a single, monolithic CML. One simplification may be to consider the possible interplay between CMLs and subsumptive robotics design ideas [20, 21, 22], focusing on egocentric leg displacement (joint angles $\alpha$ and $\beta$) independent of physical displacement.

## VI. Conclusion

Cognitive map learners (CML) are examples of ML modules, capable of being independently created then composed together to perform more complex operations. In this work, each CML controls the angular displacement of a single arm segment. Because the states are expressed as high dimensional phasor vectors, hyperdimensional computing (HDC) algebra can be used to factorize a target point, also expressed a phasor vector, into a set of arm angles without use of inverse kinematic equations. Having recovered this set of angles, the arm segment CMLs reposition themselves such that in the aggregate, the robot arm moves to the target point. This method for controlling a single arm should extend to controlling multiple arms, e.g. a quadruped robot, where a central pattern generator defines the target point per leg in time.

## Acknowledgment

Any opinions, findings and conclusions, or recommendations expressed in this material are those of the authors, and do not necessarily reflect the views of the US Government, the Department of Defense, or the Air Force Research Lab. Approved for Public Release; Distribution Unlimited: AFRL-2026-0765